\documentclass[submission,copyright,creativecommons]{eptcs}
\providecommand{\event}{Technical Communications of ICLP 2026}

\usepackage{iftex}

\ifpdf
  \usepackage{underscore}         
  \usepackage[T1]{fontenc}        
\else
  \usepackage{breakurl}           
\fi

\usepackage[utf8]{inputenc}
\usepackage{amsmath,amssymb,amsbsy,amsfonts}
\usepackage{yhmath}
\usepackage{stmaryrd}
\usepackage{microtype}
\usepackage{bm}

\newtheorem{theorem}{Theorem}
\newtheorem{proposition}{Proposition}
\newtheorem{lemma}{Lemma}

\newcommand{\eqdef}{%
  \mathrel{\vbox{\offinterlineskip\ialign{%
        \hfil##\hfil\cr%
        $\scriptscriptstyle\mathrm{def}$\cr%
        \noalign{\kern1pt}%
        $=$\cr%
        \noalign{\kern-0.1pt}%
      }}}}

\providecommand{\sysfont}{\textit}

\newcommand{\Clingo}{\sysfont{Clingo}}

\newcommand{\Clingcon}{\sysfont{Clingcon}}
\newcommand{\clingcon}{\sysfont{clingcon}}
\newcommand{\clingo}{\sysfont{clingo}}
\newcommand{\clingodl}{\clingoM{dl}}

\newcommand{\Clingodl}{\ClingoM{dl}}

\newcommand{\flingo}{\sysfont{flingo}}
\newcommand{\Flingo}{\sysfont{Flingo}}

\newcommand{\clingoM}[1]{\clingo{\small\textnormal{[}\textsc{#1}\textnormal{]}}}
\newcommand{\ClingoM}[1]{\Clingo{\small\textnormal{[}\textsc{#1}\textnormal{]}}}

\providecommand{\logfont}{\textrm}

\newcommand{\HTC}{\ensuremath{\logfont{HT}_{\!c}}}
\newcommand{\HTB}{\ensuremath{\logfont{HT}_{\!b}}}

\newcommand{\HTBSignature}{\ensuremath{\Sigma}}
\newcommand{\Sorts}{\ensuremath{\mathcal{S}}}
\newcommand{\Xbs}{\ensuremath{\mathcal{X}_s}}
\newcommand{\Dbs}{\ensuremath{\Db_s}} 
\newcommand{\orders}{\ensuremath{\mathrel{\order_s}}}
\newcommand{\ordersi}[1]{\ensuremath{\mathrel{\order_s^{#1}}}}
\newcommand{\Xb}{\ensuremath{\mathcal{X}}}
\newcommand{\Db}{\ensuremath{\mathbb{D}}} 
\DeclareMathOperator{\order}{\ensuremath{\preceq}}
\newcommand{\HTBAtoms}{\ensuremath{\mathcal{C}}}

\newcommand{\un}{\boldsymbol{u}} 

\newcommand{\val}{\ensuremath{\mathit{v}}}
\newcommand{\valb}[1]{\ensuremath{{#1}\!\downarrow}}
\newcommand{\valbs}[2]{\ensuremath{{#1\!\downarrow}_{#2}}}
\newcommand{\Hv}{\ensuremath{\mathit{h}}}
\newcommand{\Tv}{\ensuremath{\mathit{t}}}

\newcommand{\setval}{\ensuremath{\mathcal{V}}}
\newcommand{\den}[1]{\llbracket \, #1 \, \rrbracket}
\newcommand{\denHTB}[2]{\llbracket \, #1 \, \rrbracket_{#2}}

\newcommand{\domain}[1]{\ensuremath{\mathit{dom(#1)}}}
\newcommand{\vars}[1]{\ensuremath{\mathit{vars(#1)}}}

\newcommand{\cat}{\ensuremath{\mathit{c}}}

\newcommand{\fF}{\varphi}
\newcommand{\modelsht}{\models}

\newcommand{\tuple}[1]{\ensuremath{\langle #1 \rangle}}

\newcommand{\true}{\ensuremath{\mathbf{t}}}
\newcommand{\translation}{\ensuremath{\tau}}
\newcommand{\restr}[2]{\ensuremath{{#1}|_{#2}}} 

\newcommand{\id}{\ensuremath{\mathrel{\mathit{id}}}}

\newcommand{\Xc}{\ensuremath{\mathcal{X}}} 
\newcommand{\Dc}{\ensuremath{\mathbb{D}}} 
\newcommand{\HTCAtoms}{\ensuremath{\mathcal{C}}}

\newcommand{\modelshtc}{\ensuremath{\models_c}}

\newcommand{\df}[1]{\ensuremath{\mathit{def}(#1)}}

\newcommand{\code}[1]{\ensuremath{\mathtt{#1}}} 

\newcommand{\RegAtoms}{\ensuremath{\mathcal{P}}} 
\newcommand{\DcAtoms}{\ensuremath{\mathcal{D}}} 

\newcommand\diffc[3]{\code{\mathtt{\&diff}\{\mathit{#1}-\mathit{#2}\} \leq  \mathit{#3}}}

\newcommand{\anyAt}{\ensuremath{\mathtt{a}}} 
\newcommand{\anyLit}{\ensuremath{\mathtt{b}}} 
\newcommand{\vregAt}{\ensuremath{p}} 
\newcommand{\regAt}{\code{\vregAt}}
\newcommand{\dcAt}{\code{d}} 

\newcommand{\FoundedAtoms}{\ensuremath{\mathcal{F}}}
\newcommand{\ExternalAtoms}{\ensuremath{\mathcal{E}}}

\newcommand{\DC}{\ensuremath{\mathit{dc}}}
\newcommand{\XDC}{\ensuremath{\mathcal{X}}} 

\newcommand{\comp}[1]{\overline{#1}} 

\newcommand{\ext}[2]{\ensuremath{\mathit{ext}_{#2}(#1)}}
\newcommand{\orderDL}[1]{\ensuremath{#1_{\ExternalAtoms}}} 
\newcommand{\orderDLp}[2]{\ensuremath{#1_{#2}}}

\newcommand{\UnaryAtoms}{\ensuremath{\mathcal{U}}}

\newcommand{\HTBSignatureDC}{\ensuremath{\HTBSignature_{b}}} 
\newcommand{\HTCSignatureDC}{\ensuremath{\HTBSignature_{c}}}

\newcommand{\SortProp}{\ensuremath{\mathit{t}}} 
\newcommand{\SortInt}{\ensuremath{\mathit{i}}} 

\newcommand{\orderProp}{\ensuremath{\preceq_\SortProp}} 
\newcommand{\orderInt}{\ensuremath{\preceq_\SortInt}}

\newcommand{\sigmaHTC}{\ensuremath{\sigma_{\mathit{c}}}} 
\newcommand{\sigmaHTB}{\ensuremath{\sigma_{\mathit{b}}}} 

\newcommand{\HTBSignatureSystems}{\ensuremath{\HTBSignature_\star}} 

\newcommand{\SortClingcon}{\ensuremath{c}} 
\newcommand{\SortFlingo}{\ensuremath{f}} 
\newcommand{\SortDC}{\ensuremath{d}} 

\newcommand{\orderClingcon}{\ensuremath{\order_\SortClingcon}}
\newcommand{\orderFlingo}{\ensuremath{\order_\SortFlingo}}
\newcommand{\orderDC}{\ensuremath{\order_\SortDC}}

\title{Bound-Founded Semantics for Answer Set Programming with Difference Constraints: Preliminary Report}
\author{Pedro Cabalar
\institute{University of A Coruña, Spain}
\and
Jorge Fandinno
\institute{University of Nebraska at Omaha, USA}
\and
Nicolas R\"uhling
\institute{University of Potsdam, Germany}
\and
Torsten Schaub
\institute{University of Potsdam, Germany \\ Potassco Solutions, Germany}
\and
Sebastian Schellhorn
\institute{University of Potsdam, Germany}
\and
Philipp Wanko
\institute{University of Potsdam, Germany \\ Potassco Solutions, Germany}
}

\newcommand{\titlerunning}{Bound-Founded Semantics for ASP with Difference Constraints}
\newcommand{\authorrunning}{Cabalar et al.}

\hypersetup{
  bookmarksnumbered,
  pdftitle    = {\titlerunning},
  pdfauthor   = {\authorrunning},
  pdfsubject  = {EPTCS},               
  pdfkeywords = {Answer Set Programming,Difference Constraints,Bound-founded Logic of Here-and-There} 
}

\begin{document}

\maketitle

\begin{abstract}
  While the integration of linear constraints has significantly expanded the reach of Answer Set Programming (ASP),
  existing hybrid solvers often rely on disparate semantic underpinnings that lack a unified logical foundation.
  We address this gap by introducing a many-sorted variant of the Bound-founded Logic of Here-and-There (\HTB),
  providing a versatile framework capable of characterizing equilibrium models
  across a wide spectrum of alternative semantics for extensions of ASP with linear constraints.
  We apply this framework to the setting of difference constraints,
  focusing on the semantic characterization of \clingodl.
  Central to our approach is the formalization of foundedness for numeric variables.
  By investigating how different hybrid systems---such as \clingodl, \clingcon, and \flingo---justify constraint atoms,
  we uncover the semantic roots of their varying behaviors.
  This investigation results in a single,
  consistent framework that not only formalizes the foundations of current systems like \clingodl\
  but also facilitates the rigorous study of program simplifications and the future integration of diverse semantic principles.
\end{abstract}

\section{Introduction}\label{sec:introduction}

The integration of Answer Set Programming (ASP;~\cite{lifschitz19a}) with external theories,
such as linear or difference constraints~\cite{jakaosscscwa17a},
has significantly expanded its range of application,
eg.\ \cite{frscscsiwa18a,abjoossctowa21a,eigemuoeskst21a,alelge23a}.
A prominent example is the ASP system \clingodl,
which computes stable models for logic programs incorporating difference constraints.
To handle the typically infinite number of (integer) solutions,
\clingodl\ relies on a two-step algorithmic approach:
it first computes a solution and
then minimizes the integer assignments using a shortest-path algorithm~\cite{cotmal06a}.
While this approach is computationally efficient,
it has lacked a unified logical characterization within a single semantic framework.

In this paper,
we address this gap by introducing a many-sorted variant of the
Bound-founded Logic of Here-and-There (\HTB;~\cite{cafascsc19a}),
along with its nonmonotonic extension for characterizing equilibrium models
(cf.~\cite{pearce96a}).
Our framework extends the standard \HTB\ logic by incorporating many-sorted signatures to
differentiate between propositional and integer sorts,
equipped with arbitrary ordered domains.
This many-sorted approach is not only ideally suited for hybrid settings
but also provides a versatile framework for characterizing
a wide spectrum of alternative semantics within a single, uniform setting.
Most notably,
it enables us to characterize the outcomes of \clingodl's two-step computation as
equilibrium models of an \HTB\ theory.
Moreover, our framework facilitates the rigorous study of program simplifications and
the exploration of alternative semantics for hybrid logic programs.
Finally, the flexibility of our many-sorted framework allows for
the seamless integration of diverse semantic approaches.

For instance,
a pivotal semantic concept in ASP is \emph{foundedness}~\cite{gerosc91a}.
In fact, its modeling methodology relies on the ability to differentiate between founded and unfounded propositional atoms,
a distinction typically managed via choice rules.
As hybrid ASP systems incorporating numeric constraints have evolved,
diverse interpretations of foundedness have emerged
to address how theory atoms and variables are justified.
To illustrate these differences,
let us compare \clingodl\ with the \clingo\ extensions \clingcon~\cite{bakaossc16a} and \flingo~\cite{cafascwa26a}.
All three systems permit rules of the following form:%
\footnote{In fact, \clingcon\ and \flingo\ deal generally with more expressive linear constraints.}
\begin{align*}
    a\vee\neg a \qquad a\to {x>7}
\end{align*}
The resulting models vary significantly across the three systems:
\Clingcon\ produces infinitely many solutions:
one type where $a$ is true and $x$ takes in turn all values greater than 7,
and another where $a$ is false and $x$ takes all possible integer values.
\Flingo\ yields a single solution that makes $a$ false and leaves $x$ undefined,
while the solutions making $a$ true match those of \clingcon.
Finally, \clingodl\ obtains only two solutions:
one where $a$ is true and $x$ assigned $8$ and
another where $a$ is false and $x$ is undefined.
These differences are explained by their varying degrees of foundedness.
In ASP,
all true atoms must be founded, meaning they are derivable via rules from facts.
A similar principle applies to \clingodl\ and \flingo:
a variable like $x$ takes an integer value only if its containing constraint atom $x>7$ is derivable.
If the constraint atom is unfounded, $x$ remains undefined.
In contrast, \clingcon\ disconnects constrained atoms from foundedness,
treating the second rule as equivalent to $a\wedge\neg{(x>7)}\to\bot$.
\Clingodl\ further strengthens the concept of foundedness seen in \flingo\ by extending it to include the ordering of integers.
Specifically,
if $x>7$ is derivable,
\flingo\   sanctions all values greater than 7, whereas
\clingodl\ permits only the smallest admissible value, 8.
The utility of \clingodl's approach is most evident when evaluating standard difference constraints, such as
\(
x-y\leq k
\).
Such constraints are satisfied by an infinite set of integer pairs where the difference between $x$ and $y$ is at most $k$.
Rather than attempting to represent this entire infinite set, a ``canonical'' approach is typically preferred,
identifying only the minimal assignments among these pairs to represent a solution.

Just as in traditional ASP,
each of these three approaches offers significant value for modeling.
Each provides distinct mechanisms for justifying theory atoms,
enabling users to select the semantics that best aligns with their specific problem domain.
While our primary objective is to formalize the semantic foundations of \clingodl,
our bound-founded approach not only captures the underlying semantics of alternative systems, such as \clingcon\ and \flingo,
but also leverages its many-sorted framework to integrate these diverse semantics within a single, uniform setting.

The rest of the paper is structured as follows:
We start in Section~\ref{sec:htb} by introducing a many-sorted variant of \HTB.
Then,
Section~\ref{sec:htc} shows an embedding of the Logic of Here-and-There with Constraints (\HTC;~\cite{cakaossc16a}) into \HTB.
Notably, this allows us to use previous results about \clingcon~\cite{cafascwa23a,cafascwa26b} and \flingo~\cite{cafascwa26a} in the context of \HTB.
Section~\ref{sec:dc:programs} continues to introduce the syntax and semantics of
ASP with difference constraints. 
Section~\ref{sec:clingodl:htb} contains the main result of this paper,
a semantics for a subset of the language of \clingodl\ in terms of \HTB\ equilibrium models
and proposes a new semantics for ASP with difference constraints.
Finally, Section~\ref{sec:systems} unites all previous results by
showing how the three aforementioned systems can be embedded in a unifying framework in
our many-sorted \HTB\ approach.
Section~\ref{sec:conclusion} closes the paper with a brief summary and outlook.
%


\section{Many-sorted Bound-founded Logic of Here-and-There}\label{sec:htb}

We present below a many-sorted variant of \HTB, or in full detail, the Bound-founded Logic of Here-and-There,
and elaborate on its properties.

We define the many-sorted language of \HTB%
\footnote{For simplicity, we refrain from using a new acronym for this many-sorted variant of \HTB.}
over a signature
\[
  \HTBSignature = \tuple{\Sorts, {(\Xbs)_{s\in\Sorts}}, {(\Dbs, \orders)_{s\in\Sorts}}, \HTBAtoms} \, , \
  \text{ where }
\]
\begin{enumerate}
  \item $\Sorts$ is a set of sorts,
  \item ${(\Xbs)_{s\in\Sorts}}$ is a partition of a set of variables,
  \item ${(\Dbs, \orders)_{s\in\Sorts}}$ is a collection of (partially) ordered non-empty domains, and
  \item $\HTBAtoms$ is a set of constraint atoms over $\bigcup_{s\in\Sorts} \Xbs$ and $\bigcup_{s \in \Sorts} \Dbs$.
\end{enumerate}
That is,
each $\orders$ is reflexive, anti-symmetric, and transitive.
In what follows, we often use \HTBSignature\ only and leave its constituents implicit.
For convenience,
we let $\Xb$ and $\Db$ stand for $\bigcup_{s\in\Sorts} \Xbs$ and $\bigcup_{s \in \Sorts} \Dbs$, respectively.
Also, for simplicity, we associate each element of \Db\ with its representing constant.
The specific syntax of constraint atoms in \HTBAtoms\ is left open but is assumed to refer to elements of \Xb\ and \Db.
Thus,
an atom can be understood to hold or not once all variables in it are substituted by domain elements.

We define a (partial) \emph{valuation} over \HTBSignature\ as a relation
\(
\val \subseteq \Xb \times \Db
\),
such that
\begin{enumerate}
  \item if $(x,d) \in \val$ and $(x,d') \in \val$, then $d = d'$ for all $x \in \Xb$, $d,d' \in \Db$, and
  \item if $(x,d) \in \val$, then $(x,d) \in \Xbs \times \Dbs$ for some $s\in\Sorts$.
\end{enumerate}
The first condition makes sure that $\val$ behaves functionally, while
the second ensures that it respects sort information.
Since a valuation $\val$ behaves functionally,
we can also write $\val(x) = d$ if $(x,d) \in \val$
and $\val(x) = \un$ otherwise,
where $\un$
is a special symbol used to denote \emph{undefined} and different from all domain elements.
We let $\setval_{\HTBSignature}$ stand for the set of valuations over \HTBSignature\
but we drop the subscript and just write $\setval$,
whenever clear from context.
We let $\domain{\val} = \{ x \mid (x,d) \in \val\}$ denote the set of variables defined in valuation $\val$.

We define the \emph{downward closure} of a valuation \val\ over signature \HTBSignature\
as
\begin{align*}
  \valbs{\val}{\HTBSignature} = \{(x,d) \mid (x,c)\in\val, x\in\Xbs, d\in\Db_s, d \order_s c \} \, .
\end{align*}
The purpose of this closure is to enable the comparison of valuations in terms of set inclusion.
For instance, the downward closure of one valuation is strictly contained in that of another,
if the valuation assigns smaller or equal values to every variable, with strict inequality holding for at least one variable.
When clear from context, we write $\valb{v}$ instead of $\valbs{v}{\HTBSignature}$.

A formula over signature \HTBSignature\ is defined as
\begin{align*}
  \fF ::= \bot \mid \cat\in\HTBAtoms \mid \fF \land \fF \mid \fF \lor \fF \mid \fF \to \fF \, .
\end{align*}
We define $\top$ as $\bot \to \bot$ and $\neg \fF$ as $\fF \to \bot$ for any formula $\fF$.
A theory is a set of formulas.

Satisfaction of constraint atoms in \HTBAtoms\ is defined wrt \emph{denotations} over \HTBSignature\ which are
functions ${\denHTB{\cdot}{\HTBSignature}:\HTBAtoms \rightarrow 2^{\setval_{\HTBSignature}}}$
mapping atoms to sets of valuations.
Again, we drop the subscript and just write $\den{\cdot}$ whenever clear from context.
We define an interpretation over \HTBSignature\ as a pair $\tuple{\Hv,\Tv}$ of valuations over \HTBSignature\ such that
\(
\valbs{\Hv}{\HTBSignature}\subseteq\valbs{\Tv}{\HTBSignature}
\).
%
Now that all key concepts have been enriched with sorts,
the satisfaction of formulas in \HTB\ is defined as follows:
%
Let $\tuple{\Hv,\Tv}$ be an interpretation and $\fF$ a formula over \HTBSignature,
and let $\denHTB{\cdot}{\HTBSignature}$ be a denotation over \HTBSignature.
Then, $\tuple{\Hv,\Tv}$ satisfies $\fF$, written $\tuple{\Hv,\Tv}\modelsht\fF$, if the following holds:
\begin{enumerate}
  \item $\tuple{\Hv,\Tv} \not\modelsht \bot $
  \item $\tuple{\Hv,\Tv} \modelsht \cat $ iff $\val \in \denHTB{\cat}{\HTBSignature}$
        for atom $\cat\in\HTBAtoms$ and for each $\val\in\{\Hv,\Tv\}$
  \item $\tuple{\Hv,\Tv} \modelsht \fF_1\wedge\fF_2 $ iff $\tuple{\Hv,\Tv} \modelsht \fF_1$ and $\tuple{\Hv,\Tv} \modelsht \fF_2$
  \item $\tuple{\Hv,\Tv} \modelsht \fF_1\vee\fF_2 $ iff $\tuple{\Hv,\Tv} \modelsht \fF_1$ or $\tuple{\Hv,\Tv} \modelsht \fF_2$
  \item $\tuple{\Hv,\Tv} \modelsht \fF_1\rightarrow\fF_2 $ iff $\tuple{\val,\Tv} \not\modelsht \fF_1$ or $\tuple{\val,\Tv} \modelsht \fF_2$
        for each $\val\in\{\Hv,\Tv\}$
\end{enumerate}
%
We call $\tuple{\Hv,\Tv}$ a model of a theory $\Gamma$, if $\tuple{\Hv,\Tv} \modelsht \fF$ for all $\fF$ in $\Gamma$.

An interpretation $\tuple{\Tv,\Tv}$ over \HTBSignature\
is an \emph{equilibrium model} of a theory $\Gamma$ over \HTBSignature,
if
$\tuple{\Tv,\Tv} \modelsht \Gamma$ and there is no valuation $h$ over \HTBSignature\
such that $\valbs{\Hv}{\HTBSignature} \subset \valbs{\Tv}{\HTBSignature}$ and $\tuple{\Hv,\Tv} \modelsht \Gamma$.
%
If $\tuple{\Tv,\Tv}$ is an equilibrium model of $\Gamma$ over \HTBSignature,
then we call $\Tv$ a \emph{\HTBSignature\nobreakdash-stable model} of $\Gamma$.
Two theories $\Gamma_1$ and $\Gamma_2$ over \HTBSignature\
are strongly equivalent wrt \HTBSignature\ 
if $\Gamma_1\cup\Gamma$ and $\Gamma_2\cup\Gamma$ have the same
\HTBSignature\nobreakdash-stable models for any theory $\Gamma$ over \HTBSignature.

Let us illustrate how equilibrium models depend on the chosen order in \HTB.
%
Consider the theory $\Gamma = \{x\geq 1\}$ and denotation
\(
\den{x\geq 1}=\{\val\mid\val(x)\in\mathbb{Z}, \val(x)\geq 1\}
\)
over the single-sorted signature
\begin{align}\label{ex:htb:order:one}
  \HTBSignature_1 = {\tuple{\{s_1\}, \{x\}, (\{1,2\},\{ (1,1),(2,2) \}), \{x\geq 1,x\geq2\}}} \, .
\end{align}
The $\HTBSignature_1$-stable models of $\Gamma$ are
$t_1 = \{(x,1)\}$ and $t_2 = \{(x,2)\}$.
Their downward closures are $\valbs{t_i}{\HTBSignature_1} = t_i$ for $i \in \{1,2\}$.
Thus, in both cases the only possible valuation with a smaller downward closure is $\emptyset$.
Given that ${\tuple{\emptyset,t_i} \not\modelsht \Gamma}$ for ${i \in \{1,2\}}$,
we get that both $t_1$ and $t_2$ are stable models.
However, when considering 
\begin{align}\label{ex:htb:order:two}
  \HTBSignature_2 = {\tuple{\{s_2\}, \{x\}, (\{1,2\},\leq), \{x\geq 1,x\geq2\}}},
\end{align}
obtained from $\HTBSignature_1$ by using the ``lesser or equal'' relation,
the second stable model $t_2$ disappears and we only obtain $t_1$ as $\HTBSignature_2$-stable model.
This is because we now get that
\[
  {\{(x,1)\} = \valbs{t_1}{\HTBSignature_2} \subset \valbs{t_2}{\HTBSignature_2} = \{(x,1),(x,2)\}}
  \text{ and }
  {\tuple{t_1,t_2} \modelsht \Gamma}.
\]
We obtain both $\HTBSignature_1$-stable models under $\HTBSignature_2$
by adding the formula $(x \geq 2) \vee \neg (x \geq 2)$ to $\Gamma$
because $\tuple{t_1,t_2} \not\models (x \geq 2) \vee \neg (x \geq 2)$

The example illustrates that a stronger order, having more comparable domain elements,
induces a stronger minimization and thus yields fewer stable models.
Conversely, stable models are preserved when reducing the order.
The same applies to strong equivalence.
%
\begin{proposition}\label{prop:htb:order}
  Let $\HTBSignature_i = \tuple{\Sorts, {(\Xbs)_{s\in\Sorts}}, {(\Dbs, \ordersi{i})_{s\in\Sorts}}, \HTBAtoms}$
  for $i \in \{1,2\}$ be two signatures
  such that
  $\ordersi{2}$ is an extension of $\ordersi{1}$
  for all $s\in\Sorts$, and
  let $\Gamma$ and $\Gamma'$ be theories over $\HTBSignature_1$ (or $\HTBSignature_2$).
  \begin{enumerate}
    \item \label{prop:htb:order:eq-models}
          If $\Tv$ is a $\HTBSignature_2$-stable model of $\Gamma$,
          then $\Tv$ is a $\HTBSignature_1$-stable model of $\Gamma$.
    \item \label{prop:htb:order:strong-eq}
          If $\Gamma$ and $\Gamma'$ are strongly equivalent wrt $\HTBSignature_2$,
          \\
          then $\Gamma$ and $\Gamma'$ are strongly equivalent wrt $\HTBSignature_1$.
  \end{enumerate}
\end{proposition}

Next, we show that the many-sorted semantics of \HTB\ can be reduced to a single-sorted one
whenever all ordered domains behave the same on common domain elements.
%
To this end,
we define the projection of an order $\orders$ onto a set $D\subseteq\Db$ as
\begin{align*}
  \restr{\orders}{D} = {\{ (d,d') \mid (d,d') \in {\orders} \text{ and } d,d' \in D \}} \, .
\end{align*}
Two ordered domains $(\Db_1,\order_1)$ and $(\Db_2,\order_2)$ are \emph{compatible},
when
\(
\restr{\order_1}{\Db_1 \cap \Db_2} = \restr{\order_2}{\Db_1 \cap \Db_2}
\).
This 
ensures that the orders have the same behavior on common domain elements,
as well as that the union of the two orders is reflexive and anti-symmetric.
%
%
We say that sorts $s_1$ and $s_2$ are compatible,
if the ordered domains $(\Db_{s_1},\order_{s_1})$ and $(\Db_{s_2},\order_{s_2})$ are compatible.
%
%
Given a relation $R$,
we denote by $R^+$ the smallest, transitive relation containing $R$.
%
\begin{lemma}\label{lem:htb:join-sorts:order}
  Let
  \(
  \HTBSignature = \tuple{\Sorts, {(\Xbs)_{s\in\Sorts}}, {(\Dbs, \orders)_{s\in\Sorts}}, \HTBAtoms}
  \)
  be a signature with pairwise compatible sorts in $\Sorts$.
  Then, the order $\order_{s'} = (\bigcup_{s\in\Sorts} \orders)^+$ is reflexive, anti-symmetric, and transitive.
\end{lemma}
%
Furthermore,
we need to introduce additional axioms to ensure that
variables in the new single-sorted signature respect the domains of the original signature.
For this, we use two specific constraint atoms~\cite{cafascwa26b}.
Given a signature \HTBSignature,
we define the constraint atom $x : \Dbs'$
for a variable $x\in\Xbs$ and a set $\Dbs' \subseteq \Dbs$ of domain elements
with denotation
\(
\denHTB{x : \Dbs'}{\HTBSignature} = \{ v \in \setval_{\HTBSignature} \mid (x,d') \in \val, d' \in \Dbs' \}
\).
Hence, ${x : \Dbs'}$ asserts that $x$ has some value in subdomain $\Dbs'$.
This is extended to the overall domain $\Db$ by using \df{x} to assert that $x$ is defined, that is,
\(
\denHTB{\df{x}}{\HTBSignature} = \{ v \in \setval_{\HTBSignature} \mid (x,d) \in \val, d \in \Db \}
\).
%
\begin{proposition}\label{prop:htb:join-sorts}
  Let
  \(
  \HTBSignature = \tuple{\Sorts, {(\Xbs)_{s\in\Sorts}}, {(\Dbs, \orders)_{s\in\Sorts}}, \HTBAtoms}
  \)
  be a signature with pairwise compatible sorts in $\Sorts$, and
  %
  let the corresponding single-sorted signature be
  $\HTBSignature' = \tuple{\{s'\},\bigcup_{s\in\Sorts} \Xbs,(\Db_{s'},\order_{s'}),\HTBAtoms\cup\HTBAtoms_\mathit{df}}$,
  where
  ${\Db_{s'} = \bigcup_{s\in\Sorts} \Dbs}$, $\order_{s'} = (\bigcup_{s\in\Sorts} \orders)^+$
  for sort $s'\notin\Sorts$, and
  $\HTBAtoms_\mathit{df}=\bigcup_{s\in\Sorts}\{\df{x},{x:\Dbs} \mid x\in\Xbs\}$.

  Let $\Gamma$ be a theory over $\HTBSignature$ and let
  \(
  {\Gamma' = \Gamma \cup \bigcup_{s\in\Sorts} \{ \df{x} \to x : \Dbs \mid x \in \Xbs \}}
  \).

  Then, $\Tv$ is a \HTBSignature-stable model of $\Gamma$
  iff $\Tv$ is a $\HTBSignature'$-stable model of $\Gamma'$.
\end{proposition}
%
Note that expressions \df{x} and $x : \Db_{s'}$ have the same denotation for signature $\HTBSignature'$.
%

\section{Embedding \HTC\ into many-sorted \HTB}\label{sec:htc}

%
In \HTC, a signature consists of a triple $\tuple{\Xc,\Dc,\HTCAtoms}$, where
\Xc\ is a set of variables,
\Dc\ is a non-empty domain, and
\HTCAtoms\ is a set of constraint atoms.
Valuations and denotations are defined as in \HTB\ when confined to a single sort;
similarly, the satisfaction relation in \HTC, written $\modelshtc$, is analogous to \HTB.%
\footnote{We refer the reader to~\cite{cafascwa26b} for the complete definitions.}
The main difference of \HTC\ to (many-sorted) \HTB\ is that its domain is unordered (and unsorted).
Accordingly, for defining interpretations and equilibrium models,
valuations need merely be compared in terms of their degree of undefinedness,
which can be accomplished by plain set inclusion.
Hence, in \HTC,
an interpretation over $\tuple{\Xc,\Dc,\HTCAtoms}$ is a pair $\tuple{h,t}$ of valuations over $\tuple{\Xc,\Dc,\HTCAtoms}$
such that $h \subseteq t$.
An interpretation $\tuple{\Tv,\Tv}$ over $\tuple{\Xc,\Dc,\HTCAtoms}$
is an equilibrium model of a theory $\Gamma$ in \HTC\ if
$\tuple{\Tv,\Tv} \modelshtc \Gamma$ and there is no $\Hv \subset \Tv$ such that $\tuple{\Hv,\Tv} \modelshtc \Gamma$.
As above, we call $\Tv$ a stable model of $\Gamma$ in \HTC,
if $\tuple{\Tv,\Tv}$ is an equilibrium model of $\Gamma$ in \HTC.
As illustrated in~(\ref{ex:htb:order:one}--\ref{ex:htb:order:two}) and
made precise in Proposition~\ref{prop:htb:order},
stable models and strong equivalence in \HTB\ are preserved in \HTC\ as the latter employs a weaker minimization on interpretations.

We show below that any theory in \HTC\ can be expressed as a single-sorted theory in \HTB\
such that their equilibrium models coincide.
For any signature~$\tuple{\Xc,\Dc,\HTCAtoms}$ in \HTC,
we define a single-sorted signature
\(
{\HTBSignature_{\tuple{\Xc,\Dc,\HTCAtoms}} = \tuple{\{s\},\Xbs,(\Dbs,\id_s),\HTCAtoms}}
\)
in \HTB\ with
$\Xbs = \Xc$,
$\Dbs = \Dc$, and
$\id_s = \{ (d,d) \mid d \in \Dbs \}$.
Note that any formula in \HTC\ over $\tuple{\Xc,\Dc,\HTCAtoms}$ is also
a formula in \HTB\ over $\HTBSignature_{\tuple{\Xc,\Dc,\HTCAtoms}}$, and vice versa,
and thus we can consider the same theory $\Gamma$ over both signatures.
%
\begin{theorem}\label{thm:htc-htb}
  Let~$\tuple{\Xc,\Dc,\HTCAtoms}$ be an \HTC\nobreakdash-signature, and
  let
  \(
  {\HTBSignature_{\tuple{\Xc,\Dc,\HTCAtoms}} = \tuple{\{s\},\Xbs,(\Dbs,\id_s),\HTCAtoms}}
  \)
  be the corresponding \HTB-signature
  with~$\Xbs = \Xc$ and~$\Dbs = \Dc$.
  Let $\Gamma$ be a theory over $\tuple{\Xc,\Dc,\HTCAtoms}$. 

  Then,
  $\Tv$ is a stable model of $\Gamma$ in \HTC\ iff
  $\Tv$ is a ${\HTBSignature_{\tuple{\Xc,\Dc,\HTCAtoms}}}$-stable model of $\Gamma$ in \HTB.
\end{theorem}

In view of Theorem~\ref{thm:htc-htb},
we can also define a fragment of many-sorted \HTB\ that corresponds to a many-sorted version of \HTC.
The semantics of many-sorted \HTC\ is then captured in \HTB\ by the signature
\(
\tuple{\Sorts, (\Xb)_{s \in \Sorts}, (\Dbs,\id_s)_{s \in \Sorts}, \HTBAtoms}
\).
For such signatures, valuations are equal to their downward closures.
%
\begin{proposition}\label{lem:dc:id}
  Let $\val \in \setval_{\HTBSignature}$ be a valuation over an \HTB\nobreakdash-signature
  \(
  \HTBSignature = \tuple{\Sorts, (\Xb)_{s \in \Sorts}, (\Dbs,\id_s)_{s \in \Sorts}, \HTBAtoms}
  \).

  Then, we have that ${\valb{\val}} = \val$.
\end{proposition}
%

\section{Logic programs with difference constraints}\label{sec:dc:programs}

We elaborate upon the logic foundations of the extension of ASP with difference constraints~\cite{colerist09a},
and show how alternative semantics can be captured by means of many-sorted \HTB.
To this end,
we first describe the syntax of this extension~\cite{jakaosscscwa17a} and then
the semantics given in~\cite{cafascwa23a,cafascwa26b}.

\paragraph{Syntax.}
Given a set \XDC\ of integer variables over $\mathbb{Z}$,
we consider a bipartite alphabet $\langle\RegAtoms,\DcAtoms\rangle$,
consisting of disjoint
sets $\RegAtoms$ and $\DcAtoms$
of \emph{propositional} and \emph{difference constraint atoms}
(or \emph{p-} or \emph{\DC-atoms} for short, respectively).
As in \clingodl,
we denote \DC-atoms in $\DcAtoms$ by $\diffc{x}{y}{d}$
where $x,y\in\XDC$ and $d\in\mathbb{Z}$.
Furthermore,
we use function $\comp{\,\cdot\,}$ to indicate the complement of \DC-atoms, that is,
$\comp{\diffc{x}{y}{d}} = \diffc{y}{x}{-d-1}$.
The set of variables occurring in a \DC-atom is defined as $\vars{\diffc{x}{y}{d}} = \{x,y\}$.

A \emph{literal} over $\langle \RegAtoms, \DcAtoms \rangle$ and \XDC\
is an atom $\anyAt\in\RegAtoms\cup\DcAtoms$
possibly preceded by one or two occurrences of negation $\neg$.
A \emph{program} over $\langle\RegAtoms,\DcAtoms\rangle$ and \XDC\ is a set of \emph{rules} of the form
\begin{align}\label{def:dc:rule}
  \anyAt & \gets \anyLit_1,\dots,\anyLit_n
\end{align}
where
each $\anyLit_i$ is a literal over $\langle\RegAtoms,\DcAtoms\rangle$ and \XDC\ for $1 \leq i \leq n$ and
$\anyAt \in \RegAtoms\cup\DcAtoms\cup\{\bot\}$
with $\bot \not\in \RegAtoms \cup \DcAtoms$ denoting the falsum constant.
Given a rule $r$ as in~\eqref{def:dc:rule},
we refer to its \emph{head} $\anyAt$ as $h(r)$ and
let $B(r)$ be the set of atoms occurring in its \emph{body} $\anyLit_1,\dots,\anyLit_n$.
We extend this to programs in the straightforward way, that is,
$B(P) = \bigcup_{r \in P} B(r)$ and $H(P) = \{ h(r) \mid r \in P \} \setminus \{\bot\}$ for a program $P$.

\paragraph{Semantics.}
For defining semantics of logic programs with theories,
\cite{cafascwa23a} partition the set \DcAtoms\ of theory atoms%
\footnote{\DC-atoms are a special form of theory atoms.}
into two disjoint sets,
\ExternalAtoms\ and \FoundedAtoms,
standing for \emph{external} and \emph{founded} atoms, respectively.
Intuitively, the truth of an external atom requires no justification,
whereas
founded atoms must be derived through the program.
Usually,
the set \ExternalAtoms\ of external atoms consists of all theory atoms occurring in bodies,
and all other theory atoms are founded.
That is,
given a program $P$,
we have that $\ExternalAtoms = B(P) \cap \DcAtoms$
and $\FoundedAtoms = \DcAtoms \setminus \ExternalAtoms$.
Then, \cite{cafascwa26b} define a semantics of programs with theories via a translation into \HTC\ theories,
which we instantiate below for difference constraints.

Given an alphabet $\langle\RegAtoms,\DcAtoms\rangle$ and a set \XDC\ of integer variables,
we consider the \HTC\nobreakdash-signature 
\begin{align}\label{def:signature:dc:htc}
  \HTCSignatureDC & =\tuple{\Xb_\SortProp\cup\Xb_\SortInt,\{\true\}\cup\mathbb{Z},\HTBAtoms_\RegAtoms\cup\HTBAtoms_\DcAtoms\cup\HTBAtoms_\mathit{df}} \, ,
\end{align}
with
$\Xb_\SortProp=\{\vregAt \mid \regAt \in \RegAtoms\}$,%
\footnote{We use different fonts for atoms in $\langle\RegAtoms,\DcAtoms\rangle$ and their corresponding variables in \HTC\ and \HTB.}
$\Xb_\SortInt=\XDC$,
$\HTBAtoms_\RegAtoms = \{\vregAt=\true\mid \vregAt\in\Xb_\SortProp\}$,
$\HTBAtoms_\DcAtoms = \{x-y\leq d\mid x,y\in\Xb_\SortInt,d\in\mathbb{Z}\}$, and
$\HTBAtoms_\mathit{df} = \{\df{\vregAt}\mid \vregAt\in\Xb_\SortProp\}\cup\{\df{x},{x:\mathbb{Z}}\mid x\in\Xb_\SortInt\}$.
The conceptual distinction between $\SortProp$ and $\SortInt$ serves as
a precursor to our many-\linebreak[0]sorted approach in~\eqref{def:signature:dc}.
Analogously,
we define $\vars{x-y\leq d} = \{x,y\}$.
For simplicity,
we sometimes abuse notation and keep writing $\regAt$ instead of $\vregAt=\true$.
Given that the domain contains only one truth value, \true,
variables can only be true or undefined (but not false).

We use function \translation\ to map falsum, \emph{p}-, and \DC-atoms
into the \HTC\nobreakdash-signature \HTCSignatureDC\ in~\eqref{def:signature:dc:htc}:
\begin{align*}
  \translation(\bot)            & \eqdef \bot                                                       \\
  \translation(\regAt)          & \eqdef \vregAt=\true &  & \text{ for }\regAt         \in\RegAtoms \\
  \translation(\diffc{x}{y}{d}) & \eqdef x-y\leq d     &  & \text{ for }\diffc{x}{y}{d}\in\DcAtoms
\end{align*}
We require that $\vars{\dcAt} = \vars{\translation(\dcAt)}$ for each \DC-atom $\dcAt \in \DcAtoms$.
In our simplifying notation,
each \emph{p}-atom $\regAt$ is mapped into itself.
Additionally, we define
$\translation(\neg    \anyAt)=\neg    \translation(\anyAt)$ and
$\translation(\neg\neg\anyAt)=\neg\neg\translation(\anyAt)$ for
any $\anyAt\in\RegAtoms\cup\DcAtoms$.
The translation of a program $P$ is the theory $\translation(P)$ containing the implication
\begin{align}
  \translation(\anyLit_1)\wedge\dots\wedge\translation(\anyLit_n) & \to \translation(\anyAt)\label{eq:rule:translation}
\end{align}
for each rule $r \in P$ of form \eqref{def:dc:rule}.
The full translation of a program $P$ over $\langle \RegAtoms,\DcAtoms \rangle$ and \XDC\ into \HTC\
is done relative to a set $\ExternalAtoms\subseteq\DcAtoms$ of external atoms~\cite{cafascwa26b}:
\begin{align}
  \translation_c(P,\RegAtoms,\XDC,\ExternalAtoms) & = \translation(P)\cup\delta(\RegAtoms,\XDC)\cup\sigmaHTC(\ExternalAtoms) \qquad\text{ with}\label{eq:translation:htc} \\
  \delta(\RegAtoms,\XDC)                          & = \{ \df{\vregAt}\to\vregAt=\true     \mid \regAt\in\RegAtoms \} \cup
  \{ \df{x}      \to     x :\mathbb{Z}\mid x \in \XDC \}                                   \label{eq:delta}                                                               \\
  \sigmaHTC(\ExternalAtoms)                       & = \{                     \df{x}\mid x \in \vars{\dcAt}, \dcAt \in \ExternalAtoms \} \label{eq:sigma:htc} \, .
\end{align}
Intuitively, $\delta(\RegAtoms,\XDC)$ forces propositional and integer variables to range only over their corresponding domains.
In fact,
\df{\vregAt} and \df{x} have the same denotation as $\vregAt:{\{\true\}\cup\mathbb{Z}}$ and $x:{\{\true\}\cup\mathbb{Z}}$.
The atoms in $\sigmaHTC(\ExternalAtoms)$ ensure that variables in external atoms are defined.

Following the characterization in~\cite{cafascwa26b},
we define the \emph{stable models} of a program $P$ over $\langle \RegAtoms,\DcAtoms \rangle$ and \XDC\
wrt a set $\ExternalAtoms\subseteq\DcAtoms$ of external atoms as the
\HTCSignatureDC-stable models of the theory $\translation_c(P,\RegAtoms,\XDC,\ExternalAtoms)$ in \HTC.
Note that any such stable model satisfies one of $\dcAt$ or $\comp{\dcAt}$
for all external \DC-atoms $\dcAt\in\ExternalAtoms$~\cite{cafascwa26b}.

For illustration, consider the program consisting of the rules:
\begin{align}
  \diffc{x}{y}{-1} & \label{ex:diff:one:one}                        \\
  \regAt           & \gets \diffc{x}{y}{-2} \label{ex:diff:one:two}
\end{align}
For the set $\{ \diffc{x}{y}{-2} \}$ of external atoms,
we obtain infinitely many stable models of the following two types, namely,
\begin{align}
  \val(\vregAt)=  \un   & \text{ and } \val(x), \val(y) \in \mathbb{Z}, \val(x)-\val(y)   =  -1\text{, and}\label{ex:stable:models:one} \\
  \val(\vregAt)=\,\true & \text{ and } \val(x), \val(y) \in \mathbb{Z}, \val(x)-\val(y) \leq -2.           \label{ex:stable:models:two}
\end{align}

\paragraph{Semantics of~\protect\clingodl.}
The ASP system \clingodl\ computes the stable models of logic programs incorporating difference constraints.
The system relies on a two-step algorithm for computing a canonical set of stable models.
First, a stable model is computed by treating \DC-atoms in rule bodies as external.
Second, the integer part of the computed model is then minimized using a shortest-path algorithm~\cite{cotmal06a}.
In our example in~(\ref{ex:diff:one:one}--\ref{ex:diff:one:two}),
this approach yields just two canonical stable models,
which represent the two infinite classes of stable models in \eqref{ex:stable:models:one} and \eqref{ex:stable:models:two}.

To capture these canonical models in our setting,%
\footnote{The correspondence of this order to the implementation of \clingodl~(v5) is posited by its developers.}
we define a partial order $\orderDL{\leq}$
between valuations
$\val$, $\val'$ relative to a set $\ExternalAtoms$ of (external) \DC-atoms as follows:
\(
\val \orderDL{\leq} \val'
\)
iff
\begin{enumerate}
  \item $\domain{\val} = \domain{\val'}$,
  \item $\ext{\val}{\ExternalAtoms} = \ext{\val'}{\ExternalAtoms}$, and
  \item $\val(x)\leq \val'(x) \text{ for all } x\in \domain{\val}\cap\Xb_\SortInt$.
\end{enumerate}
where
\(
\ext{\val}{\ExternalAtoms} = \{ \dcAt \in \ExternalAtoms \mid \val \in \den{\translation(\dcAt)} \}
\)
is the set of atoms in \ExternalAtoms\ satisfied by $\val$.
As usual, we write $\val \orderDL{<} \val'$ whenever $\val \orderDL{\leq} \val'$ and $\val \neq \val'$.
For example, we get $\{(\vregAt=\true),(x,0)\}\orderDL{<}\{(\vregAt=\true),(x,2)\}$,
while the following stable models are incomparable:
\begin{enumerate}
  \item $\{(\vregAt=\true),(x,0)\}$ and $\{(x,0)\}$       \hfill (different propositional variables)
  \item $\{(x,0)\}$                 and $\{(y,0)\}$       \hfill (different integer variables)
  \item $\{(x,0),(y,1)\}$           and $\{(x,1),(y,0)\}$ \hfill (incomparable valuations)
\end{enumerate}
Intuitively,
this order compares stable models satisfying the same set of atoms
and ranks them according to the values assigned to their integer variables.
To see this,
note that the first condition along with the singleton domain of $\Xb_\SortProp$ implies that
$\val(x) = \val'(x) \text{ for all } x\in \domain{\val}\cap\Xb_\SortProp$,
thus, $\val$ and $\val'$ satisfy the same $p$-atoms.
The second condition ensures that $\val$ and $\val'$ satisfy the same external \DC-atoms.
Given that only $p$- and external \DC-atoms occur in rule bodies,
it follows that $\val$ and $\val'$ sanction the same set of head atoms,
thus, effectively satisfying the same overall set of atoms.

We call a valuation \val\ non-negative, if $\val(x) \geq 0$ for all $x \in \domain{\val}\cap\Xb_\SortInt$.

Then, a non-negative valuation $t$ is a \clingodl\nobreakdash-stable model of a program $P$ if
\begin{enumerate}
  \item $t$ is a stable model of $P$ wrt $B(P) \cap \DcAtoms$
  \item there is no non-negative valuation \val\ such that
        \begin{enumerate}
          \item $\val\orderDLp{<}{B(P) \cap \DcAtoms}t$ and
          \item \val\ is a stable model of $P$ wrt $B(P) \cap \DcAtoms$.
        \end{enumerate}
\end{enumerate}

We obtain two \clingodl\nobreakdash-stable models
for our example program in~(\ref{ex:diff:one:one}--\ref{ex:diff:one:two}), namely,
\begin{align}
  \{                (x,0),(y,1)\} & \text{ and }\label{ex:clingodl:stable:model:one} \\
  \{(\vregAt,\true),(x,0),(y,2)\} & \, .        \label{ex:clingodl:stable:model:two}
\end{align}

For programs without external atoms,
the definition of the order simplifies to the first and third conditions above.
This matches the proposal by~\cite{cafascwa23a},
which considers only programs where \DC-atoms are restricted to rule heads.
%

\section{Logic programs with difference constraints in \HTB}\label{sec:clingodl:htb}

We now present translations of programs with \DC-atoms into \HTB\ theories.

Given an alphabet $\langle\RegAtoms,\DcAtoms\rangle$ of $p$- and \DC-atoms and
a set \XDC\ of integer variables,
we consider the \HTB\nobreakdash-signature \HTBSignatureDC\ defined as
\begin{align}\label{def:signature:dc}
      \HTBSignatureDC & =\tuple{\{\SortProp,\SortInt\},(\Xb_\SortProp,\Xb_\SortInt), ((\{\true\},\orderProp),(\mathbb{Z},\orderInt)),\HTBAtoms_\RegAtoms\cup\HTBAtoms_\DcAtoms\cup\HTBAtoms_\mathit{df}\cup\HTBAtoms_\UnaryAtoms}
\end{align}
where $\SortProp$ and $\SortInt$ are truth and integer sorts;
$\Xb_\SortProp$, $\Xb_\SortInt$, $\HTBAtoms_\RegAtoms$, $\HTBAtoms_\DcAtoms$, and $\HTBAtoms_\mathit{df}$ are defined as in~\eqref{def:signature:dc:htc};
and $(\{\true\},\orderProp)$ and $(\mathbb{Z},\orderInt)$ are the ordered domains
for sorts $\SortProp$ and $\SortInt$, respectively,
where
$\orderProp$ is the singleton order $\true\orderProp\true$ over $\{\true\}$
and
$\orderInt$ represents the usual order $\leq$ on $\mathbb{Z}$.
Furthermore, $\HTBAtoms_\UnaryAtoms = \{x\circ d \mid x \in \Xb_\SortInt,d\in\mathbb{Z},\circ\in\{\leq,\geq\}\}$.

Notably, we show that
the order among valuations in \HTB\
(leveraging order $\orderInt$ together with additional axioms)
allows us to capture the formation of canonical models in \clingodl\ presented above.
To this end,
we keep using function \translation\ to map programs into theories,
but now over \HTB-signature \HTBSignatureDC\ rather than \HTC-signature \HTCSignatureDC\ in~\eqref{def:signature:dc}.
The full translation of a program $P$ over $\langle \RegAtoms,\DcAtoms \rangle$ and \XDC\ into \HTB\
wrt a set $\ExternalAtoms\subseteq\DcAtoms$ of external atoms
is given as
\begin{align}
      \translation_b(P,\XDC,\ExternalAtoms) & = \translation(P) \cup \gamma(\XDC) \cup \sigmaHTB(\ExternalAtoms) \qquad\text{ with}    \label{eq:translation:htb} \\
      \gamma(\XDC)                          & = \{ x : \mathbb{Z} \to x \geq 0 \mid x \in \XDC \}                                      \label{eq:gamma}           \\
      \sigmaHTB(\ExternalAtoms)             & = \{ \translation(\dcAt) \vee \translation(\comp{\dcAt}) \mid \dcAt \in \ExternalAtoms \}\label{eq:sigma:htb}
\end{align}
The axioms in $\gamma(\XDC)$ ensure that integer variables do not take on negative values,
as required for \clingodl-stable models.
Notably, those in $\sigmaHTB(\ExternalAtoms)$ are stronger than the axioms in $\sigmaHTC(\ExternalAtoms)$ presented in~\eqref{eq:sigma:htc} for \HTC.
To ensure that we only compare stable models satisfying the same set of external atoms
(cf.\ definition of $\orderDL{\leq}$),
we require in~\eqref{eq:sigma:htb} that either the translation of an external atom or its complement is satisfied.
This requirement cannot be met by simply stipulating that the integer variables appearing in external \DC-atoms are defined,
as done in \eqref{eq:sigma:htc},
because the minimization inherent to \HTB\ would then assign these variables their minimum possible values without regard for the specific bounds imposed by the external constraints.
In contrast, the axioms $\delta(\RegAtoms,\XDC)$ in \eqref{eq:delta} are now 
unnecessary by our use of the many-sorted signature \HTBSignatureDC.

For example,
the translation $\translation_b(\{\eqref{ex:diff:one:one},\eqref{ex:diff:one:two}\},\{x,y\},\{\diffc{x}{y}{-2}\})$ yields
\begin{align}
      x-y \leq -1                     \label{ex:dc:one:one:htb}   \\
      x-y \leq -2    \to \regAt       \label{ex:dc:one:two:htb}   \\
      x : \mathbb{Z} \to x \geq 0     \label{ex:dc:one:gamma:one} \\
      y : \mathbb{Z} \to y \geq 0     \label{ex:dc:one:gamma:two} \\
      (x-y \leq -2) \vee (y-x \leq 1) \label{ex:dc:one:sigma:direct}
\end{align}
This theory has two \HTBSignatureDC-stable models which correspond to the \clingodl-stable models of
the program in (\ref{ex:diff:one:one}--\ref{ex:diff:one:two}),
given in (\ref{ex:clingodl:stable:model:one}--\ref{ex:clingodl:stable:model:two}), namely,
\begin{align*}
      \{          (x,0),(y,1)\}       & \text{ and } \\
      \{(\vregAt,\true),(x,0),(y,2)\} & \, .
\end{align*}

To understand why the simpler axioms $\sigmaHTC(\ExternalAtoms)$ from \HTC\ are insufficient
for our many-sorted \HTB\ translation,
consider the interpretation
\(
\tuple{\Hv,\Tv}=
\tuple{\{(x,0),(y,1)\},\{(\vregAt,\true),(x,0),(y,2)\}}
\).
While \tuple{\Hv,\Tv} satisfies the formulas in (\ref{ex:dc:one:one:htb}--\ref{ex:dc:one:gamma:two}),
it fails to satisfy the disjunctive axiom~\eqref{ex:dc:one:sigma:direct}.
This failure occurs because
${\Hv\not\in\den{x-y \leq -2}}$ and ${\Tv\not\in\den{y-x \leq 1}}$,
a condition that effectively ``protects'' $t$ as a \HTBSignatureDC-stable model.
However, if we were to use $\sigmaHTC(\ExternalAtoms)$ instead,
we would obtain $\tuple{\Hv,\Tv}\models\df{x}$ and $\tuple{\Hv,\Tv}\models\df{y}$.
Consequently, $\Tv$ would no longer be a \HTBSignatureDC-stable model
because it is ``defeated'' by the smaller valuation $h$.

Furthermore,
the axioms in $\sigmaHTB(\ExternalAtoms)$ constitute a stronger version of the law of excluded middle
by requiring $\translation(\dcAt)\vee\translation(\comp{\dcAt})$ rather than $\translation(\dcAt)\vee\neg\translation(\dcAt)$
for external \DC-atoms \dcAt.
This distinction is exemplified by
the single rule \eqref{ex:diff:one:two}
and its two \clingodl-stable models:
$\{(\vregAt,\true),(x,0),(y,2)\}$ and $\{(x,0),(y,0)\}$.
While the \HTB\ translation using the stronger axiom captures both models,
replacing~\eqref{ex:dc:one:sigma:direct} with $(x-y \leq -2) \vee \neg(x-y \leq -2)$
causes the second model to disappear.
This occurs because default negation ($\neg$) allows variables to remain undefined.
In particular, the interpretation $\tuple{\emptyset,\{(x,0),(y,0)\}}$ satisfies the weaker theory,
thereby ``defeating'' the valuation $\{(x,0),(y,0)\}$ as an equilibrium model.

We obtain the following general correspondence.
\begin{theorem}\label{thm:dc:main}
      Let $P$ be a program over $\langle \RegAtoms,\DcAtoms \rangle$ and \XDC,
      and $\ExternalAtoms\subseteq\DcAtoms$ a set of external atoms.

      Then,
      $t$ is a \clingodl\nobreakdash-stable model of $P$
      iff
      $t$ is a $\HTBSignatureDC$-stable model of $\translation_b(P,\XDC,\ExternalAtoms)$.
\end{theorem}
%
This allows us to define strong equivalence for programs in \clingodl\ in terms of \HTB.

%
We now focus on the case where \DC-atoms occur only in the head;
specifically, for a program $P$,
we assume $B(P)\cap\DcAtoms=\emptyset$,
implying that $\ExternalAtoms=\emptyset$ and $\FoundedAtoms=\DcAtoms$.
Under these conditions, our translation simplifies to:
$\translation_b(P,\XDC,\emptyset) = \translation(P) \cup \gamma(\XDC)$.
We show that this restriction is sufficient to characterize external \DC-atoms.
%
The idea is to translate a program $P$ with external \DC-atoms $B(P)\cap\DcAtoms$ in rule bodies
into a program in which all \DC-atoms occur in rule heads only.

To this end, let
$\RegAtoms_\DcAtoms = \{\regAt_\dcAt\mid\dcAt\in\DcAtoms\}$ be a set of fresh, propositional atoms
such that $\RegAtoms_\DcAtoms\cap\RegAtoms=\emptyset$.
Given a set $D \subseteq \DcAtoms$ of \DC-atoms,
let $Q(D)$ be the program
\begin{align}\label{eq:dc:ext}
      Q(D) & = \{
      \regAt_{\dcAt} \gets \neg\neg\regAt_{\dcAt},
      \dcAt \gets \regAt_{\dcAt},
      \comp{\dcAt} \gets \neg \regAt_{\dcAt} \mid \dcAt \in D
      \} \, .
\end{align}
For a program $P$,
let $P^*$ be the result of
replacing in each rule $r \in P$ every occurrence of
a \DC-atom $\dcAt \in B(r) \cap \DcAtoms$ in the rule body
by the corresponding propositional atom $\regAt_{\dcAt}$.

For example, the translation $P^*$ of our program $P$ in~(\ref{ex:diff:one:one}--\ref{ex:diff:one:two}) is
\begin{align}
      \diffc{x}{y}{-1} & \label{prg:translation:one}                                 \\
      \regAt           & \gets \regAt_{\diffc{x}{y}{-2}} \label{prg:translation:two}
\end{align}
together with the following rules from $Q(\{\diffc{x}{y}{-2}\})$
\begin{align}
      \regAt_{\diffc{x}{y}{-2}} & \gets \neg\neg\regAt_{\diffc{x}{y}{-2}}\label{prg:translation:tri} \\
      \diffc{x}{y}{-2}          & \gets \regAt_{\diffc{x}{y}{-2}}        \label{prg:translation:for} \\
      \diffc{y}{x}{ 1}          & \gets \neg \regAt_{\diffc{x}{y}{-2}}   \label{prg:translation:fiv}
\end{align}
Notably, in $P^*\cup Q(\{\diffc{x}{y}{-2}\})$, all rule bodies are devoid of \DC-atoms.

Applying translation \translation\ to this program
yields the \HTB\ theory containing implications of form \eqref{eq:rule:translation}
for all rules in~\eqref{prg:translation:one} to~\eqref{prg:translation:fiv}.
Note, however, that the translation of \eqref{prg:translation:tri} is equivalent to
$\translation(\regAt_{\diffc{x}{y}{-2}}) \vee \neg \translation(\regAt_{\diffc{x}{y}{-2}})$.
Hence, the rules for a \DC-atom \dcAt\ in~\eqref{eq:dc:ext} characterize the behavior of external atoms
by providing a choice between \dcAt\ and its complement $\comp{\dcAt}$,
effectively ensuring that one of the two is derived.
The resulting \HTB-theory has the same stable models as those given in
\eqref{ex:clingodl:stable:model:one} and \eqref{ex:clingodl:stable:model:two}
except that the model in \eqref{ex:clingodl:stable:model:two}
additionally satisfies the propositional atom $\vregAt_{\diffc{x}{y}{-2}}=\true$.

In general,
our translation preserves the original stable models under projection onto the initial variables
after all \DC-atoms in rule bodies are eliminated.
To this end,
let $\HTBSignatureDC^+$ be the signature obtained by
extending $\Xb_\SortProp$ and $\HTBAtoms_\RegAtoms$ in $\HTBSignatureDC$ with
$\{\vregAt_{\dcAt}\mid\regAt_\dcAt\in\RegAtoms_\DcAtoms\}$ and $\{\vregAt_{\dcAt} = \true\mid\regAt_\dcAt\in\RegAtoms_\DcAtoms\}$, respectively.
%
\begin{proposition}\label{prop:dc:external-head}
      Let $P$ be a program over $\langle \RegAtoms, \DcAtoms \rangle$ and \XDC,
      and let $\ExternalAtoms = B(P) \cap \DcAtoms$ be the set of external atoms.
      Let $t$ be a valuation over $\HTBSignatureDC$ and
      let
      \(
      t' = t \cup \{ (\vregAt_\dcAt,\true) \mid t\in\denHTB{\translation(\dcAt)}{\HTBSignatureDC}, \dcAt\in\ExternalAtoms\}
      \)
      be a valuation over $\HTBSignatureDC^+$.
      Then,
      $t$ is a $\HTBSignatureDC$-stable model of $\translation_b(P,\XDC,\ExternalAtoms)$
      iff
      $t'$ is a $\HTBSignatureDC^+$-stable model of $\translation_b(P^* \cup Q(\ExternalAtoms),\XDC,\emptyset)$.
\end{proposition}

%
So far,
we have employed translations of programs into \HTB\ to investigate their logical properties.
We now leverage this framework to provide a general semantics for ASP with difference constraints.
Our approach permits the use of \DC-atoms in 
rule heads and bodies,
treating them uniformly as founded atoms.

Given a program $P$ over $\langle \RegAtoms,\DcAtoms \rangle$ and \XDC,
we define a \emph{founded model} of $P$ as
a $\HTBSignatureDC$-stable model of $\translation(P) \cup \gamma(\XDC)$.
For illustration,
the \HTB\ theory obtained for our example program in~(\ref{ex:diff:one:one}/\ref{ex:diff:one:two}) is
\begin{align*}
      x-y \leq -1    &                 \\
      x-y \leq -2    & \to \regAt      \\
      x : \mathbb{Z} & \to x\geq0      \\
      y : \mathbb{Z} & \to y\geq0 \, .
\end{align*}
This theory has a single $\HTBSignatureDC$-stable model $\{(x,0),(y,1)\}$,
which thus constitutes the single founded model of our example program.

Recall that if $\diffc{x}{y}{-2}$ was treated as an external atom,
there would be another solution with $\{(\vregAt,\true),(x,0),(y,2)\}$.
In fact, both valuations are incomparable in terms of the order $\orderDL{\leq}$
due to different propositional variables (and different external atoms),
and hence both are returned by the current \clingodl\ system.
On the other hand, the founded model yields only the model assigning the smaller value to $y$
(and the same value for $x$),
which are the minimal, justified values by the fact in \eqref{ex:diff:one:one}.

In the example in~(\ref{ex:diff:one:one}--\ref{ex:diff:one:two}),
the founded model is also a \clingodl-stable model.
However, this does not hold in general as the following program shows.
\begin{align}
      \diffc{x}{y}{-1} & \label{ex:diff:two:one}                        \\
      \regAt           & \gets \diffc{x}{z}{-2} \label{ex:diff:two:two}
\end{align}
This program has a single founded model $\{(x,0),(y,1)\}$;
however, its \clingodl-stable models are $\{(x,0),(y,1),(z,0)\}$ and $\{(\vregAt,\true),(x,0),(y,1),(z,2)\}$.
This discrepancy arises because the founded model leaves variable $z$ undefined,
whereas $z$ is assigned a value in any \clingodl-stable model
since the new atom $\diffc{x}{z}{-2}$ is treated as external.

Notably,
for programs where \DC-atoms are restricted to rule heads,
the founded and \clingodl-stable models coincide.
This stems from the fact that such programs contain no external atoms;
in this case, the translation
$\translation_b(P,\XDC,\emptyset)$ reduces to $\translation(P) \cup \gamma(\XDC)$.
Consequently,
by employing the translation to eliminate all \DC-atoms from rule bodies,
we can characterize \clingodl-stable models in terms of founded models.

A natural question is whether founded models can be characterized as stable models via a specific ordering,
as is done for \clingodl.
This possibility, however, is refuted by the following counterexample.
Consider the program comprising the fact in~\eqref{ex:diff:one:one} along with the following two rules:
\begin{align}
      \regAt \vee \neg\regAt & \label{ex:diff:three:one}                \\
      \diffc{x}{y}{-2}       & \gets  \regAt  \label{ex:diff:three:two}
\end{align}
While this program shares the same stable models~(\ref{ex:clingodl:stable:model:one}--\ref{ex:clingodl:stable:model:two})
as the one in~(\ref{ex:diff:one:one}--\ref{ex:diff:one:two}),
it yields the two founded models $\{(x,0),(y,1)\}$ and $\{(\regAt,\true),(x,0),(y,1)\}$---%
rather than just one.
%

\section{A unifying framework for hybrid ASP systems}\label{sec:systems}

This section consolidates our findings to establish that many-sorted \HTB\ offers a versatile framework
capable of unifying the semantic underpinnings of diverse hybrid ASP solvers.
We remain within the context of logic programs with difference constraints
and consider the semantics of systems \clingcon, \flingo, and \clingodl.

To this end,
we partition
the set $\XDC=\XDC_{\SortClingcon}\cup\XDC_{\SortFlingo}\cup\XDC_{\SortDC}$ of integer variables
into three disjoint sets, each corresponding to one of the three systems,
and define
the set $\DcAtoms$ of \DC-atoms
as $\DcAtoms_{\SortClingcon}\cup\DcAtoms_{\SortFlingo}\cup\DcAtoms_{\SortDC}$
with $\DcAtoms_j = \{\dcAt\in\DcAtoms\mid\vars{\dcAt}\subseteq\XDC_j\}$ for $j\in\{\SortClingcon,\SortFlingo,\SortDC\}$.
This ensures that each \DC-atom contains only variables of the same sort,
and thus adheres to a single semantics.

As in previous sections,
we consider programs over the alphabet $\langle\RegAtoms,\DcAtoms\rangle$ and \XDC,
now refined by this tripartite division of constraints and variables.
Crucially, this tripartite division is reflected in the categorization of \DC-atoms as either external or founded.
While \clingcon\ treats all \DC-atoms as external,
\flingo\ considers them exclusively as founded.
In the case of \clingodl,
we maintain the flexibility of the variants discussed previously.
Specifically, given a program $P$,
we define
$\ExternalAtoms = ((B(P)\cup H(P))\cap\DcAtoms_\SortClingcon) \cup (B(P) \cap \DcAtoms_\SortDC)$
and
$\FoundedAtoms  = \DcAtoms_{\SortFlingo}\cup (\DcAtoms \setminus \ExternalAtoms) = \DcAtoms \setminus \ExternalAtoms$.

The enriched structure of our programs' alphabet is reflected by the signature of corresponding \HTB\ theories.
Given an alphabet $\langle\RegAtoms,\DcAtoms\rangle$ of $p$- and \DC-atoms and
a set \XDC\ of integer variables, as defined above,
we consider the \HTB-signature \HTBSignatureSystems\ defined as
\begin{align*}
  \HTBSignatureSystems & =
  \tuple{\Sorts,(\Xb_\SortProp,\Xb_\SortClingcon,\Xb_\SortFlingo,\Xb_\SortDC), ((\{\true\},\orderProp),(\mathbb{Z},\orderClingcon),(\mathbb{Z},\orderFlingo),(\mathbb{Z},\orderDC)),\HTBAtoms_\RegAtoms\cup\HTBAtoms_\DcAtoms}
  \quad\text{ where}
\end{align*}
\begin{enumerate}
  \item $\Sorts = \{\SortProp,\SortClingcon,\SortFlingo,\SortDC\}$
        for the truth sort $\SortProp$ in~\eqref{def:signature:dc}, and
        sorts $\SortClingcon$, $\SortFlingo$, and $\SortDC$ reflecting \clingcon, \flingo, and \clingodl, respectively,
  \item $\orderClingcon$, $\orderFlingo$, $\orderDC$ correspond to the relations
        $\id_\SortClingcon$, $\id_{\!\SortFlingo}$,  $\leq$ over $\mathbb{Z}$,
  \item $\HTBAtoms_\RegAtoms$ as in~\eqref{def:signature:dc} or~\eqref{def:signature:dc:htc}, respectively, and
        $\HTBAtoms_\DcAtoms = \bigcup_{j\in\{\SortClingcon,\SortFlingo,\SortDC\}}\{x-y\leq d\mid x,y\in\Xb_j,d\in\mathbb{Z}\}$.
\end{enumerate}

The unified translation of programs into \HTB-theories is built upon the mapping $\translation_b$
defined in~\eqref{eq:translation:htb}.
While the underlying mapping $\translation$ of programs from Section~\ref{sec:dc:programs} remains independent of sorts,
the non-negativity axioms in~\eqref{eq:gamma} are applied to every integer variable
for the sake of uniformity,
although, strictly speaking this would not be necessary for \clingcon\ and \flingo.
The disjunctive axioms in~\eqref{eq:sigma:htb} ensure the correct logical behavior for
external \DC-atoms in both $\DcAtoms_\SortClingcon$ and $\DcAtoms_\SortDC$.
Conversely,
\DC-atoms belonging to the \flingo\ partition $\DcAtoms_{\!\SortFlingo}$ are excluded
from these axioms due to their founded nature.
Finally,
although the weaker axioms in~\eqref{eq:sigma:htc} would suffice for \clingcon-style \DC-atoms,
we adopt a uniform translation for the sake of simplicity;
this choice is justified by the fact that the axioms in~\eqref{eq:sigma:htb} 
imply the ones in~\eqref{eq:sigma:htc}.
%
%

\section{Conclusion}
\label{sec:conclusion}

We have presented a unified logical foundation for hybrid ASP systems incorporating difference constraints
by introducing a many-sorted variant of the Bound-founded Logic of Here-and-There (\HTB).
By extending \HTB\ with many-sorted signatures,
we have provided a framework that differentiates between propositional and integer sorts
while accommodating arbitrary ordered domains.
This structure allows for a versatile characterization of equilibrium models
across a wide spectrum of alternative semantics for logic programs with linear constraints.
Our approach provides significant insight into the formation of equilibrium models in existing systems
based on their varying degrees of foundedness.
In our formalism, these differences translate into
whether a system relies upon ordered domains and
whether it treats constraint atoms as external or founded.
This reflects the core principle of stable model formation in ASP%
---where atoms remain `false' unless they are provably `true'---%
which is analogous to utilizing an ordering on Boolean values where `true' is greater than `false'~\cite{azchst13a}.

We have shown that the canonical models of \clingodl%
---typically obtained through a two-step algorithmic approach---%
correspond to equilibrium models over ordered integers in \HTB.
In contrast,
systems such as \clingcon\ and \flingo\ rely on unordered integer domains.
This is reflected by their previous embeddings~\cite{cafascwa23a} in the Logic of Here-and-There with Constraints (\HTC) and
the subsequent embedding of \HTC\ into \HTB\ presented above.
Furthermore,
\cite{cafascwa23a} show that \clingcon\ treats all constraint atoms as external,
while \flingo\ treats them exclusively as founded.

Additionally,
we elaborated on \clingodl's semantics by distinguishing between external and founded \DC-atoms.
We demonstrated that programs containing external \DC-atoms in rule bodies can be systematically reduced to a ``head-only'' form.
Beyond characterizing existing solvers,
we leveraged \HTB\ to propose a general founded semantics for ASP with difference constraints,
treating all \DC-atoms uniformly as founded.
Ultimately, the flexibility of our many-sorted approach ensures that
diverse semantic principles can be seamlessly integrated into a single, uniform logical setting.
%
%
%

\paragraph{Acknowledgments}
We would like to thank the anonymous reviewers for their valuable feedback that allowed us to improve presentation of several points.
This work was supported
by grant PID2023-148531NB-I00 funded by Spanish
Ministry MCIU/AEI/10.13039/501100011033, funds FEDER, EU,
by the NSF CAREER award 2338635, USA,
and by DFG grant SCHA 550/15, Germany.
Any opinions, findings, and conclusions or recommendations expressed in this material are those of the authors and do not necessarily reflect the views of the National Science Foundation.

\end{document}